\documentclass[10pt,twocolumn,letterpaper]{article}

\usepackage{wacv}              

\usepackage{graphicx}
\usepackage{amsmath}
\usepackage{amssymb}
\usepackage{booktabs}
\usepackage{graphicx}
\usepackage{booktabs}
\usepackage{amsmath}
\usepackage{amssymb}
\usepackage{multirow}
\usepackage{bm}
\usepackage{comment}
\usepackage{color, colortbl}
\usepackage{wrapfig}
\usepackage{ulem}
\usepackage[table,dvipsnames]{xcolor}
\usepackage{color,soul}

\definecolor{Gray}{gray}{0.9}
\newcolumntype{g}{>{\columncolor{Gray}}c}

\usepackage[pagebackref,breaklinks,colorlinks]{hyperref}

\usepackage[capitalize]{cleveref}
\crefname{section}{Sec.}{Secs.}
\Crefname{section}{Section}{Sections}
\Crefname{table}{Table}{Tables}
\crefname{table}{Tab.}{Tabs.}

\def\wacvPaperID{XXXXX} 
\def\confName{WACV}
\def\confYear{2025}

\begin{document}

\title{CrowdMAC: Masked Crowd Density Completion \\ for Robust Crowd Density Forecasting}

\author{Ryo Fujii$^{1}$ \hspace{0.6cm} Ryo Hachiuma$^{2}$  \hspace{0.6cm} Hideo Saito$^{1}$ \\
$^{1}$Keio University \hspace{0.6cm} $^{2}$NVIDIA \\
{\tt\small \{ryo.fujii0112, hs\}@keio.jp, rhachiuma@nvidia.com}\\
}

\maketitle

\begin{abstract}
A crowd density forecasting task aims to predict how the crowd density map will change in the future from observed past crowd density maps. However, the past crowd density maps are often incomplete due to the miss-detection of pedestrians, and it is crucial to develop a robust crowd density forecasting model against the miss-detection. This paper presents a MAsked crowd density Completion framework for crowd density forecasting (CrowdMAC), which is simultaneously trained to forecast future crowd density maps from partially masked past crowd density maps (\ie, forecasting maps from past maps with miss-detection) while reconstructing the masked observation maps (\ie, imputing past maps with miss-detection).
Additionally, we propose Temporal-Density-aware Masking (TDM), which non-uniformly masks tokens in the observed crowd density map, considering the sparsity of the crowd density maps and the informativeness of the subsequent frames for the forecasting task. Moreover, we introduce multi-task masking to enhance training efficiency. In the experiments, CrowdMAC achieves state-of-the-art performance on seven large-scale datasets, including SDD, ETH-UCY, inD, JRDB, VSCrowd, FDST, and croHD. We also demonstrate the robustness of the proposed method against both synthetic and realistic miss-detections. The code is released at
\href{https://fujiry0.github.io/CrowdMAC-project-page}{project page}.
\end{abstract}

\section{Introduction}
\label{sec:intro}
Accurately forecasting a crowd's movement in a video is crucial for various applications, such as advancing navigation systems to select an optimal route for agents (robots) and effectively minimizing the risk of collisions~\cite{Aroor2018AAMAS, Liu2019IROS}. A crowd's movement can be naively forecasted by applying the well-studied trajectory prediction framework~\cite{Alahi2016CVPR, Gupta2018CVPR, Salzmann2020ECCV, Mangalam2021ICCV}. Despite significant attention and numerous proposed works on the trajectory prediction task, most existing works~\cite{Alahi2016CVPR, Gupta2018CVPR, Salzmann2020ECCV, Mangalam2021ICCV} rely on the critical assumption that each trajectory is successfully obtained by two preprocessing (\ie, upstream perception) modules, pedestrian detection~\cite{Ren2015NIPS, Redmon2016CVPR, Lin2017ICCV, Nicolas2020ECCV} and tracking~\cite{Bewley2016ICIP, Wojke2017ICIP, Chen2019ICCV, Zhang2022ECCV}. However, this assumption is not feasible for real-world scenarios since the complete trajectory for each pedestrian is not always available due to the miss-detection of pedestrians or incorrect tracking. Using estimated trajectories as the input for existing trajectory predictors could lead to significant accumulated errors and pose serious threats to safety~\cite{Xu2024ICRA}.

Crowd density forecasting task~\cite{Minoura2021RAL}, which directly forecasts future crowd density maps from the past crowd density maps that are estimated from either pedestrian detection~\cite{Ren2015NIPS, Redmon2016CVPR, Lin2017ICCV, Nicolas2020ECCV} or crowd density estimation~\cite{Zhang2016CVPR, Li2018CVPR, Wang2019CVPR, Gao2020TCSVT} modules, offers the potential as an alternative approach for forecasting the movements of pedestrians in real-world robotic applications, even though the movement of \textit{each} pedestrian is not forecasted. However, the accuracy of the crowd density forecasting task still highly relies on the obtained past crowd density maps. Thus, it is crucial to develop a robust crowd density forecasting model against the error accumulated by the upstream crowd density estimation or pedestrian detection modules. (\ie, miss-detection).

The masked autoencoders (MAE)~\cite{He2022CVPR} has garnered significant attention due to its recent achievements in image-based self-supervised learning. This approach involves masking a portion of patches in the input data and reconstructing the missing patches using an autoencoder structure. Inspired by the remarkable capabilities of MAE in reconstructing masked patches (interpolation) of spatio-temporal data, we explore its potential for forecasting future patches (extrapolation) in the crowd density forecasting task. This paper presents \textit{CrowdMAC}, which forecasts future crowd maps from observed crowd maps. Unlike MAE, which masks a subset of patches during training, we mask frames (entire patches) in the future density maps, and the model is trained to reconstruct the masked future frames.

As previously mentioned, to robustly forecast the crowd density map with the miss-detection of the pedestrians, the patches in the input past observed crowd density maps are masked. During training, the model jointly reconstructs the masked patches in past observed maps and predicts the patches in future maps. Through the simultaneous training of reconstruction and future prediction, the model acquires meaningful representations that enhance its ability to predict future maps while improving its robustness against miss-detections.

In addition, drawing inspiration from representation learning approaches that utilize MAE with non-uniform masking based on token informativeness across various domain data inputs~\cite{li2022NeurIPS,Sun2023CVPR,Min2023IV,Mao2023ICCV}, we propose {\it temporal-density-aware masking} (TDM) strategy that non-uniformly masks observed maps during the training to facilitate the training and obtain better forecasting performance. The TDM comprises two components: temporal-aware ratio and density-aware masking. First, since the last observed frame is the most informative in the forecasting task compared to the first observed frame, the first module increases the masking ratio along the time index of the maps. Second, given the observation that the input crowd density maps are sparse (\eg, pedestrians may not exist in all pixels in the map) and that reconstructing these sparse pixels doesn’t aid model training, we propose to use the accumulated density as empirical semantic richness
prior to effectively guiding the crowd density map masking process, enabling enhanced attention to semantically rich spatial regions.  

Moreover, we adopt multi-task learning using different masks inspired by MegVit~\cite{Yu2023CVPR}, which addresses various video synthesis tasks with a single model using different masking schemes for efficient training. Specifically, we train the model with three tasks: future prediction, past prediction, and interpolation. Training with these tasks enables the model to establish a robust bidirectional relationship between past and future motion and serves as a form of regularization, enhancing overall performance.

We extensively benchmark the performance of the proposed model well-established trajectory prediction datasets, ETH-UCY~\cite{Pellegrini2010ECCV,Leal2014CVPR}, SDD~\cite{Robicquet2016ECCV}, inD~\cite{Bock2020IV}, and JRDB~\cite{Martin2021TPAMI} and crowd analysis datasets, FDST~\cite{Fang2019ICME}, croHD~\cite{Sundararaman2021CVPR}, and VSCrowd~\cite{Li2022TIP}. Despite the simplicity of the proposed framework, it outperforms previous crowd forecasting methods by large margins under two evaluation settings: ground truth inputs and pedestrian detection inputs. Moreover, our model surpasses conventional trajectory prediction approaches in both settings. In addition, we investigate the robustness of crowd density forecasting approaches against pedestrian miss-detection, a critical consideration in practical scenarios. The study validates that the proposed method outperforms previous approaches.

Our contributions are summarized as follows:

\begin{itemize}
\item We present a robust crowd density forecasting approach, \textit{CrowdMAC}, that simultaneously reconstructs the masked tokens in the past crowd density maps and predicts the future crowd density maps during the training.

\item We propose {\it temporal-density-aware masking} (TDM) strategy that non-uniformly masks observed maps during the training to facilitate the training and obtain better forecasting performance. 

\item We introduce a straightforward yet highly effective multi-task masking scheme that facilitates the learning of bidirectional motion connections and serves as a form of regularization to enhance performance.
\end{itemize}

\begin{figure*}[tb] 
\centering
\includegraphics[width=\linewidth]{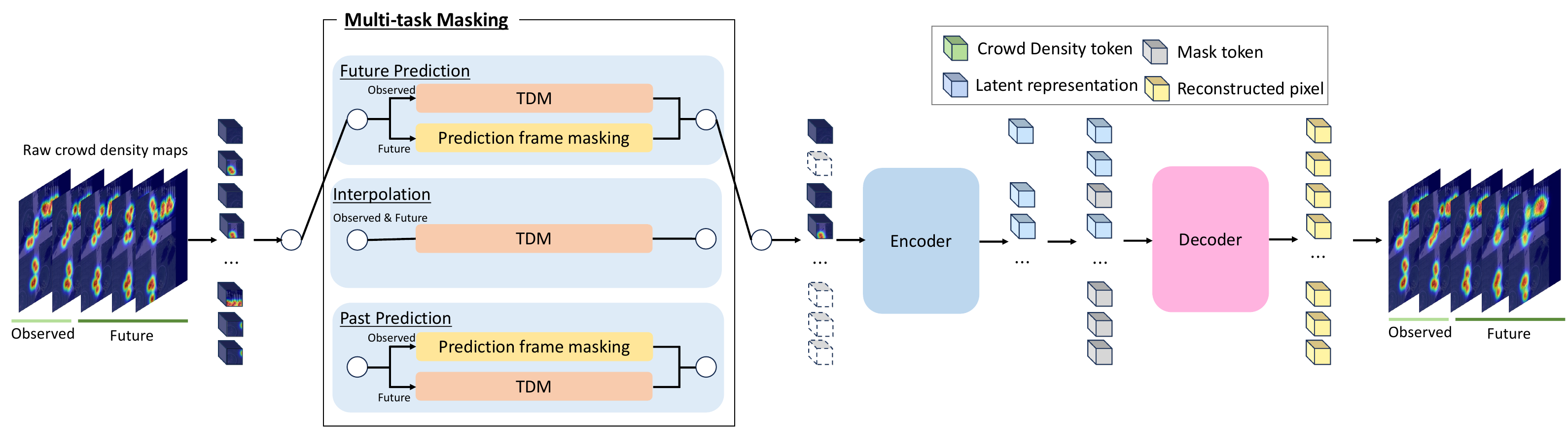}
\caption{Overview of the CrowdMAC during the training phase. Cube embedding transforms the crowd density maps into a sequence of tokens. At each training step, one of the task masks is sampled, and a subset of tokens is masked. The remaining tokens, along with the space-time position embedding, are fed into a Transformer encoder and decoder to reconstruct the masked maps. Note that the multi-task masking is applied only during the training phase, while the only future prediction task mask is used during the inference phase.}
\label{fig:overview}
\end{figure*}

\section{Related Work}
\subsection{Trajectory Prediction}
Human trajectory forecasting aims to infer plausible future positions from observed trajectories. Deep learning-based trajectory prediction approaches~\cite{Alahi2016CVPR, Gupta2018CVPR, Mohamed2020CVPR, Xu2022CVPR, Kosaraju2019Neurips, Mangalam2021ICCV, Sadeghian2019CVPR, Ivanovic2019ICCV, Mangalam2020ECCV, Salzmann2020ECCV, FUJII2021Access, Gu2022CVPR, Mao2023CVPR} have become dominant due to their impressive representational ability. Some studies focus on better modeling interactions with other agents, such as pedestrians~\cite{Alahi2016CVPR, Gupta2018CVPR, Mohamed2020CVPR, Xu2022CVPR} and the environment~\cite{Kosaraju2019Neurips, Mangalam2021ICCV} and some works~\cite{Gupta2018CVPR, Sadeghian2019CVPR, Kosaraju2019Neurips, Ivanovic2019ICCV, Mangalam2020ECCV, Salzmann2020ECCV, Gu2022CVPR, Mao2023CVPR} aim to model the multi-modality of future motions.

However, despite the increasing accuracy of trajectory prediction methods, they work on trajectory prediction in an idealized setting where ground truth past trajectories are used as inputs for both training and evaluation without considering upstream perception errors (\ie, pedestrian detection and tracking error). However, this idealized setting is not feasible in the real world since the trajectories that contain errors estimated from the upstream modules are inputted to the trajectory forecasting approaches.

\subsection{Crowd Density Forecasting}
Crowd density forecasting aims to forecast crowd density maps of future frames. PDFN-ST~\cite{Minoura2021RAL}, the patch-based crowd density forecasting model, is pioneering work in the field of crowd density forecasting. It models the diverse and complex crowd density dynamics of the scene based on spatially or spatiotemporally overlapping patches. Crowd density forecasting directly forecasts future crowd density maps from past crowd density maps. The past crowd density maps are estimated either from upstream crowd density estimation or the object detection modules. Thus, similar to the trajectory prediction method, the crowd density forecasting method highly depends on the accuracy of the upstream module. Therefore, developing a robust crowd density forecasting model against errors accumulated by the upstream modules, such as object detection, is of critical importance and serves as the motivation for this work. CrowdMAC aims to enhance robustness against the miss-detection of pedestrians.

\subsection{Masked Autoencoders (MAE)}
Masked autoencoders~\cite{He2022CVPR,Feichtenhofer2022NEURIPS, Wang2023CVPR} are a type of denoising autoencoder~\cite{Vincent2008ICML} that obtain meaningful representations by learning to reconstruct images corrupted by masking. In particular, the masked autoencoder (MAE)~\cite{He2022CVPR} approach accelerates pre-training to learn representations from images by using an asymmetric architecture. VideoMAE~\cite{Tong2022NEURIPS} directly extends MAE and efficiently reconstructs the pixels of masked video patches with a high masking ratio through tube masking. Our approach leverages the efficiency of the masked autoencoder approach and extends it to the crowd density forecasting task. The original masked autoencoder approaches aim to obtain powerful pretrained models to improve the accuracy of video-related downstream tasks. On the other hand, we focus on the interpolation capabilities of the masked autoencoder itself and apply it to the forecasting task, which is robust to missing inputs (\ie, miss-detection).

\subsection{Mask Strategy}
Various ablation experiments of MAE on various inputs have shown that the performance on downstream tasks depends on the masking strategy. Rather than randomly mask sampling in input images~\cite{He2022CVPR} or videos~\cite{Tong2022NEURIPS,Huang_2023_ICCV}, recent custom masking approaches have been proposed for various inputs such as images~\cite{li2022NeurIPS}, videos~\cite{Sun2023CVPR}, 3D LiDAR point clouds~\cite{Min2023IV}, or sequential skeletons~\cite{Mao2023ICCV}, trajectories~\cite{Chen_2023_ICCV}. These approaches non-uniformly sample tokens based on the \textit{informativeness} of each token. SemMAE~\cite{li2022NeurIPS} utilizes a masking approach based on semantically segmented parts of objects in images to obtain a powerful image representation. Traj-MAE~\cite{Chen_2023_ICCV} introduces a social-temporal masking approach, where social masking reconstructs each agent’s trajectory based on the trajectories of surrounding agents, and temporal masking focuses on reconstructing the randomly masked positions of agents. The informativeness-based masking strategy improves performance on downstream tasks such as image or video recognition. Building on these methods, we propose a temporal-density-aware masking strategy that masks informative tokens in the input crowd density maps during training, specifically for the crowd density forecasting task.

\section{CrowdMAC}
\subsection{Problem Definition}
\label{sec:problem}
The goal of crowd density forecasting is to predict future crowd density maps over $T_{pred}$ timesteps (\ie, \textit{future frames}), $C_{pred} = [c_{T_{obs}+1},\ldots,c_{T_{obs}+T_{pred}}]$ from the observed past crowd density maps over $T_{obs}$ timesteps (\ie, \textit{past frames}), $C_{obs} = [c_{1},\ldots,c_{T_{obs}}]$. $c_t\in[0,1]^{W \times H}$ represents the crowd density map at time step $t$ with a size of $(W, H)$. The observed crowd density maps $C_{obs}$ are extracted from each input RGB video frame $I_t$ using the off-the-shelf pedestrian detection~\cite{Lin2017ICCV, Nicolas2020ECCV, Han2023ICCV} or crowd density estimation~\cite{Wang2019CVPR, Gao2020TCSVT} approaches.

\begin{figure}[tb]
\centering
\begin{tabular}{ccc}
\begin{minipage}{0.3\hsize}
    \begin{center}
        \includegraphics[clip, width=\hsize]{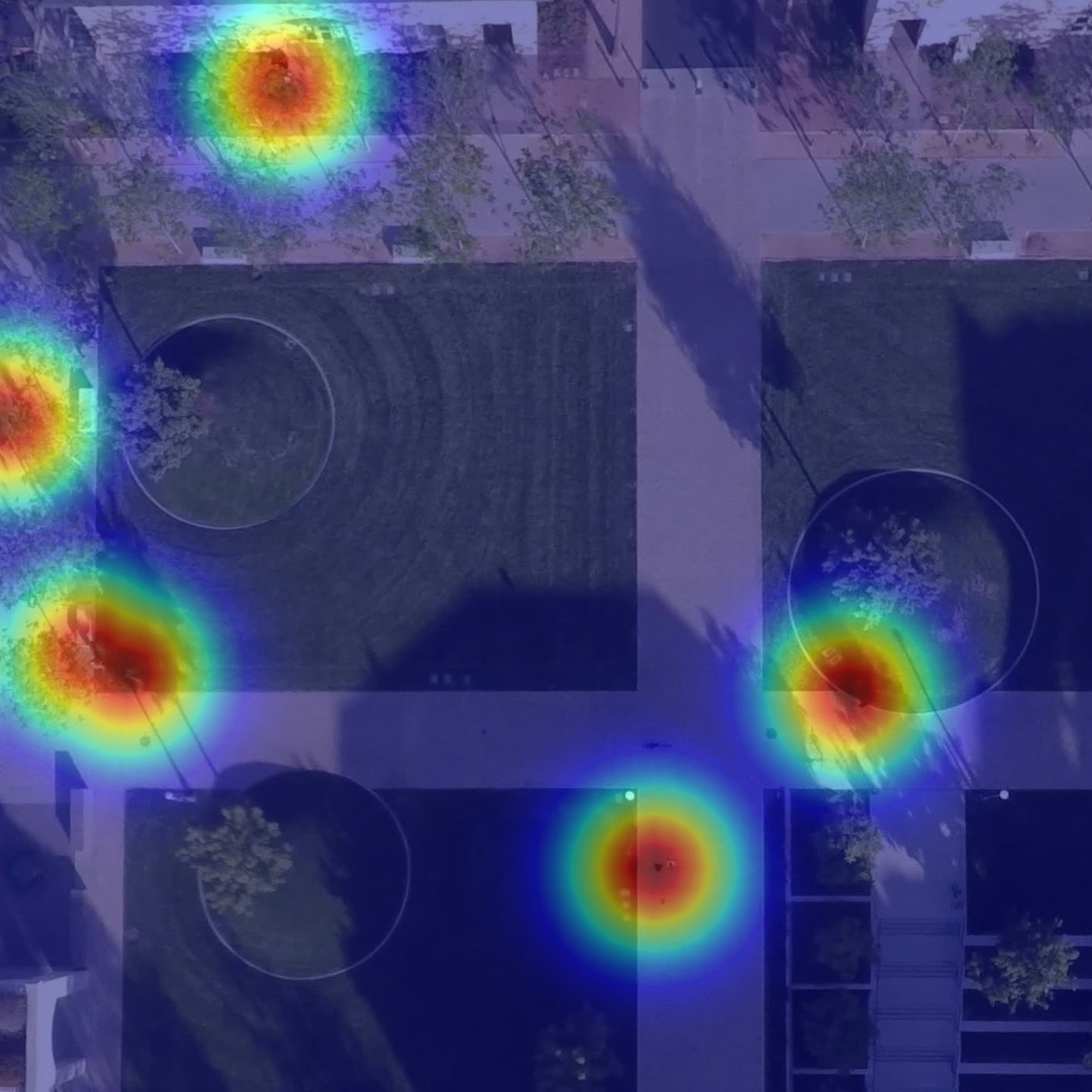} 
         {\scriptsize (a) SDD}
    \end{center}
\end{minipage}
&
\begin{minipage}{0.3\hsize}
    \begin{center}
        \includegraphics[clip, width=\hsize]{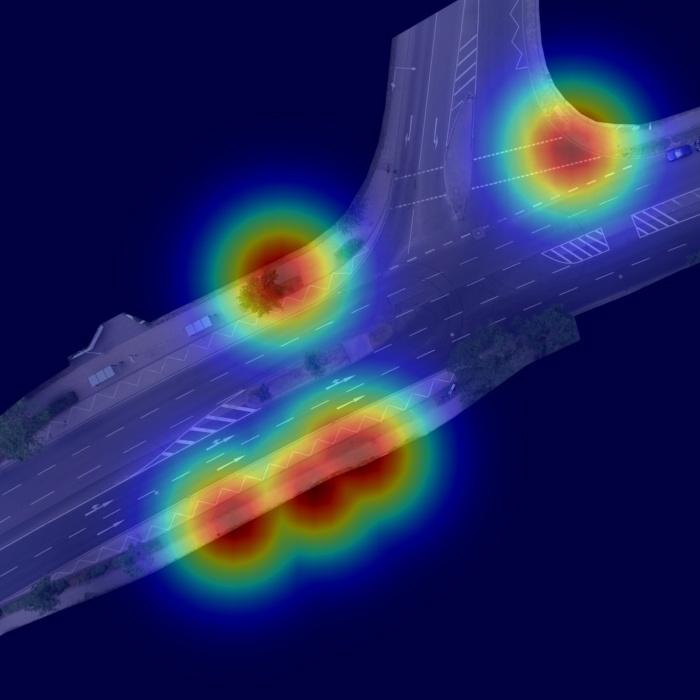} 
        {\scriptsize (b) inD}
    \end{center}
\end{minipage}
&
\begin{minipage}{0.3\hsize}
    \begin{center}
        \includegraphics[clip, width=\hsize]{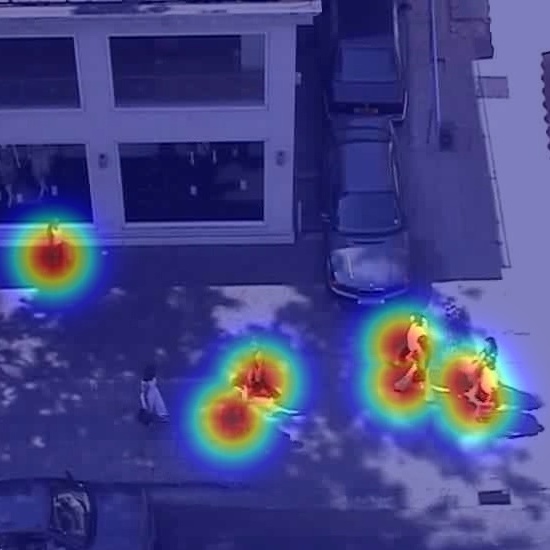}  
        {\scriptsize (c) ETH-UCY}
    \end{center}
\end{minipage}

\end{tabular}
\caption{Examples of input crowd density maps in the various trajectory prediction datasets. The density maps are visualized as a heatmap and overlayed onto the RGB image for visualization purposes.}
 \label{fig:example-sparse-heatmap}
\end{figure}

\subsection{Overview}
\cref{fig:overview} illustrates the overview of the proposed CrowdMAC. CrowdMAC takes crowd density maps $C\in\mathbb{R}^{T \times H \times W}$ as input, comprising of observed crowd density maps $C_{obs}\in\mathbb{R}^{T_{obs} \times H \times W }$ and future crowd density maps $C_{pred}\in\mathbb{R}^{T_{pred} \times H\times W}$, where $T=T_{obs}+T_{pred}$. It employs the space-time cube embedding~\cite{Tong2022NEURIPS} to transform the input maps into a set of token embeddings $X \in \mathbb{R}^{F \times N_sN_r}$, where $F$ is the channel dimension of the tokens, $N_s=HW/H_cW_c$ and $N_r=T/T_c$ are the numbers of tokens along the spatial and temporal directions, respectively. $T_c$, $H_c$, and $W_c$ are the size of each token along temporal, height, and width dimensions, respectively. Then, a subset of tokens is non-uniformly masked with a multi-task masking scheme. The remaining tokens are fed into a Transformer encoder and decoder~\cite{Vaswani2017NIPS}, along with the space-time position embedding, to reconstruct the masked maps. The loss function is the mean squared error (MSE) loss between the masked tokens and the reconstructed ones.

\subsection{Multi-Task Masking}
Instead of training the model solely on a single future prediction task, we adopt a multi-task approach involving three tasks: future prediction, past prediction, and interpolation. In the future prediction task, CrowdMAC simultaneously reconstructs partially masked tokens in the past crowd density maps and predicts the entire future crowd density maps. For the interpolation task, CrowdMAC reconstructs partially masked tokens in both the past and future crowd density maps. In the past prediction task, CrowdMAC reconstructs the partially masked tokens in the future crowd density maps while predicting the entire past crowd density maps. At each training step, one of these tasks is randomly sampled. This multi-task training approach enables the model to establish a robust bidirectional relationship between past and future motion. It also serves as a regularization technique, reducing overfitting by requiring the model to perform well across multiple objectives, leading to more generalized learning.

For future and past prediction tasks, we apply the proposed temporal-density-aware masking (TDM) along with frame masking to select tokens for masking. In the interpolation task, only TDM is used to select the masked tokens.

\begin{figure}[tb]
\begin{center}
\resizebox{\columnwidth}{!}{
\begin{minipage}{\hsize}
  \begin{center}
    \includegraphics[clip, width=\hsize]{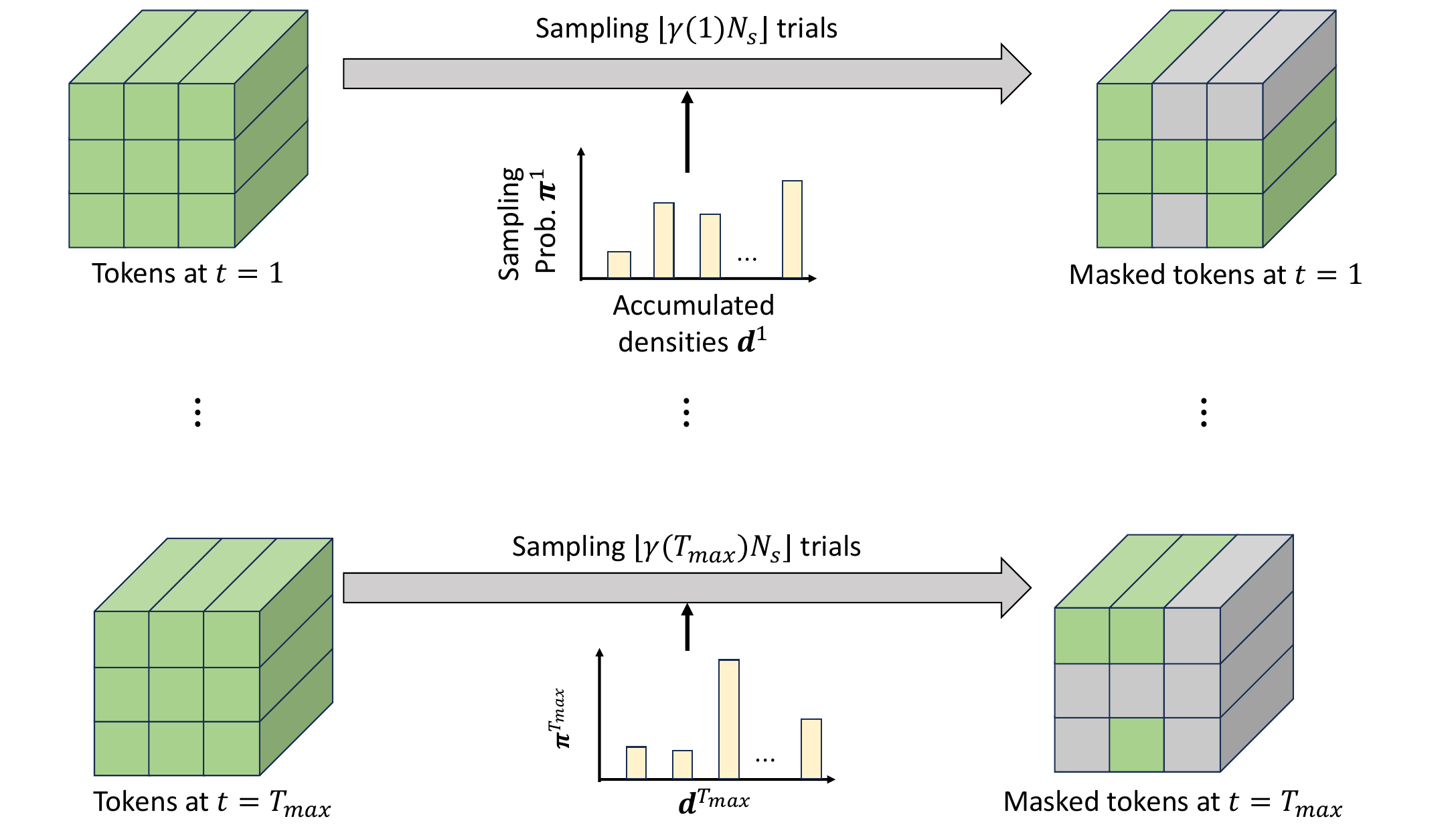} 
  \end{center}
\end{minipage}}

\caption{The conceptual figure of the proposed Temporal-density-aware masking (TDM) module. The number of masked tokens is monotonically increased along with the time step (Temporal-aware Masking Ratio). For each time index $t$ of the tokens, the tokens are non-uniformly selected based on the probability of the accumulated density values of tokens (Density-aware Masking). After the masking, the tokens colored in gray indicate that the tokens are masked.}
\label{fig:tdm}
\end{center}
\end{figure}

\subsection{Temporal-Density-aware Masking (TDM)}
Instead of uniformly masking tokens in frames via random sampling with a fixed masking ratio, like MAE or VideoMAE, during the training, we propose Temporal-Density-aware Masking (TDM) that randomly masks informative tokens. The overview of TDM is visualized in~\cref{fig:tdm}. TDM consists of two modules: temporal-aware masking (TM) ratio and density-aware masking (DM). TM calculates the number of masked tokens for each time step from the tokens' time index, and the sampling probability of each token in each time step is calculated from DM based on the accumulated density value of each token. The details of each module are explained in the following sections.

\subsubsection{Temporal-aware Masking (TM) Ratio}
For the future prediction task and interpolation, we assume the latter frame is more informative than the first frame. Therefore, instead of sampling tokens randomly with a fixed masking ratio $\gamma$ that computes the number of the tokens to be masked~\cite{Tong2022NEURIPS,Chen_2023_ICCV}, the proposed TM ratio $\gamma(t)$ is a monotonically increasing function of the tokens' time index of the observation tokens $t \in \{1,\ldots, T_{max}\}$, where $T_{max}$ is set to $T_{obs}/T_c$ for the future prediction task and $T/T_c$ for the interpolation task. By designing $\gamma(t)$ to increase along with the time index, the model is trained to reconstruct more informative tokens in the subsequent frames.
Concretely, we design the $\gamma(t)$ as follows:

\begin{equation}
\gamma(t) = 1 - e^{(-\lambda t / T_{max})},
\label{eq:ratio}
\end{equation}
where $\lambda$ is sampled from the uniform distribution $U[0, \lambda_{max}]$ every epoch with hyper-parameter $\lambda_{max}$. 

For the past prediction task, we assume the earlier frame is more informative than the last
frame.  The TM ratio $\gamma(t)$ is a monotonically decreasing function of the tokens' time index of the future tokens $t \in \{1,\ldots, T_{max}\}$, where $T_{max}$ is set to $T_{pred}/T_c$. We design the $\gamma(t)$ as follows:

\begin{equation}
\gamma(t) = 1 - e^{(-\lambda (T_{max}-t) / T_{max})}.
\end{equation}

\subsubsection{Density-aware Masking (DM)}
In contrast to RGB images or videos, crowd density maps are often sparse, meaning that pedestrians do not exist in all the pixels in the maps, and most pixel values are empirically 0. For example, \cref{fig:example-sparse-heatmap} visualizes the estimated crowd density maps from RGB images in the various datasets (\eg SDD~\cite{Robicquet2016ECCV}, inD~\cite{Bock2020IV}, and ETH-UCY~\cite{Pellegrini2010ECCV,Leal2014CVPR}). We observe that most of the tokens are sparse\footnote{Statistically, $9.6\%$, $5.6\%$, $8.8\%$ pixels in the map have density values higher than the particular threshold (0.1) on SDD, inD, and ETH-UCY datasets, respectively.} (the accumulated pixel density values of patches mostly, being 0), and training to reconstruct these tokens through MAE is not effective~\cite{Sun2023CVPR,Mao2023ICCV} (the patch lacks \textit{informativeness} during the training). Therefore, we propose to non-uniformly sample the tokens based on the accumulated density value of each token.

Specifically, we first compute the accumulated density value $d_i$ for each token ${\bm x}_i \in X$ by taking a sum of the density values in the pixels that belong to the token as follows:
\begin{equation}
    d_i = \sum_{j \in \Omega_i} C_j,
\end{equation}
where $\Omega_i$ denotes the set of pixels that belong in the $i$-th token.
Then, we calculate the masking probability vector ${\bm \pi}^t \in \mathbb{R}^{N_s}$ for each token's time index using the softmax function as follows:
\begin{equation}
    {\bm \pi}^t = {\rm Softmax}({\bm d}^t/\tau),
\label{eq:prob}
\end{equation}
where ${\bm d}^t \in \mathbb{R}^{N_s}$ is a vector of accumulated densities for each time index $t$, and $\tau$ is a hyper-parameter that controls the sharpness of the softmax function.
Finally, the indices of the masked tokens in the observation frames are obtained by sampling a Multinomial distribution according to the probabilities ${\bm \pi}^t$ for $\lfloor \gamma(t)N_s \rfloor$ trials without replacement for every time index $t$.

\section{Experiments}
\subsection{Datasets}
We use a total of seven datasets to investigate the performance of CrowdMAC. Specifically, we conduct experiments on four datasets for trajectory prediction: ETH-UCY~\cite{Pellegrini2010ECCV,Leal2014CVPR}, Stanford Drone Dataset (SDD)~\cite{Robicquet2016ECCV}, Intersection Drone Dataset (InD)\cite{Bock2020IV}, and JackRabbot Dataset (JRDB)~\cite{Martin2021TPAMI}. Furthermore, we evaluate our approach on three datasets for crowd analysis tasks (\eg, crowd detection, localization, and counting), including Fudan-ShanghaiTech (FDST)~\cite{Fang2019ICME}, Crowd of Heads Dataset (CroHD)~\cite{Sundararaman2021CVPR}, and surveillance-view Video
Crowd dataset (VSCrowd)~\cite{Li2022TIP}. Since the CroHD dataset only provides the detection results from the off-the-shelf object detector, we employ these instead of manually annotated detection results in the experiments.

\subsection{Implementation Details}
 We employ the ViT-Small~\cite{Alexey2021ICLR} backbone. Following prior work~\cite{Minoura2021RAL}, we use the input size of $80\times80$ for $20$ frames which consists of $8$ past frames ($T_{obs}$) and $12$ future frames ($T_{pred}$) in \cref{sec:problem}. Note that the $8$ past frames are used as input during the inference. We set the space-time cube size for the embedding along the spatial dimensions as $W_c=8$, $H_c=8$, and along the temporal dimension as $T_c=4$. 
 We employ the mean squared error (MSE) loss between the masked pixels and the reconstructed pixels for training. We train our model with AdamW optimizer~\cite{Ilya2018ICLR} with a base learning rate $5e^{-4}$ and weight decay $1e^{-5}$ for $1200$ epochs. We train the model from scratch for SDD and use the pretrained model from SDD for initialization when training on other datasets. The hyperparameters were determined through a standard coarse-to-fine grid search or step-by-step tuning. We warm up training for $60$ epochs, starting from learning rate $1e^{-6}$, and decay it to $0$ throughout training using cosine decay. We set the batch size to $256$ and train the model using two NVIDIA TITAN RTX GPUs. We employ spatial data augmentation techniques, such as rotation, scaling, horizontal flip, and vertical flip of the crowd density maps. We empirically set $\lambda_{max}$ (\cref{eq:ratio}) of temporal-aware masking as $9$. Also, we set $\tau$ in density-aware masking (\cref{eq:prob}) to $500$.

\subsection{Evaluation Metrics}
Followed by prior work~\cite{Minoura2021RAL}, we use Jensen-Shannon (JS) divergence to measure the performance of the forecasting.
We report the Average JS divergence ($AD_{JS}$) and Final JS divergence ($FD_{JS}$). $AD_{JS}$ is the divergence between the predicted and ground truth map averaged over all the future time steps, while $FD_{JS}$ is the divergence between the predicted and ground truth map at the final time step.

\subsection{Comparison Models} \label{comparison-models}
Given that only one previous work has addressed the crowd density forecasting task, apart from the crowd density forecasting method PDFN-ST (RA-L'21)~\cite{Minoura2021RAL}, we adopt standard and state-of-the-art trajectory prediction approaches, Y-Net (ICCV'21)~\cite{Mangalam2021ICCV} and Social-Transmotion (ICLR'24)~\cite{Saadatnejad2024ICLR}, as comparison models, following the previous approach~\cite{Minoura2021RAL}. To evaluate trajectory prediction models using the crowd density forecasting metric, we generate crowd density maps from the predicted trajectories. Similar to the creation of ground truth crowd density maps, we apply a 2D Gaussian filter to the predicted positions of each pedestrian in each frame.

\subsection{Evaluation Protocols} 
\label{sec:eval_protoc}
We evaluate the efficacy of our method under a common setting in the state-of-the-art trajectory prediction model, which involves observing trajectories over 8-time steps (equivalent to 3.2 seconds) and subsequently predicting future trajectories spanning the next 12 time steps (equivalent to 4.8 seconds), employing two distinct protocols.

\begin{table}[tb]
\caption{Comparison of the crowd density forecasting and trajectory prediction approaches using ground truth pedestrian positions (see \cref{sec:eval_protoc}). The lower metrics ($AD_{JS}, FD_{JS}$) are better. As FDST and croHD do not provide annotations of person identity for the test set, we do not report the results of the trajectory prediction methods for these datasets. The results of each ETH-UCY subset are provided in the supplementary.}

\begin{center}
\resizebox{\columnwidth}{!}{
\begin{tabular}{cccccccgg}
\toprule
 \multirow{4}{*}{Dataset} & \multicolumn{4}{c}{Trajectory Prediction} & \multicolumn{4}{c}{Crowd Density Forecasting}  \\ 
 \cmidrule(l{2pt}r{3pt}){2-5} 
 \cmidrule(l{2pt}r{3pt}){6-9} 
&\multicolumn{2}{c}{Y-Net~\cite{Mangalam2021ICCV}}   &\multicolumn{2}{c}{Social-Trans.~\cite{Saadatnejad2024ICLR}} &\multicolumn{2}{c}{PDFN-ST~\cite{Minoura2021RAL}} & \multicolumn{2}{g}{Ours}  \\
 \cmidrule(l{2pt}r{3pt}){2-3} \cmidrule(l{2pt}r{3pt}){4-5}
 \cmidrule(l{2pt}r{3pt}){6-7} \cmidrule(l{2pt}r{3pt}){8-9} 
 & $AD_{JS}$ & $FD_{JS}$ & $AD_{JS}$ & $FD_{JS}$ & $AD_{JS}$ & $FD_{JS}$ & $AD_{JS}$ & $FD_{JS}$ \\
\toprule
 SDD~\cite{Robicquet2016ECCV} &  0.098 & 0.144 & 0.083 & \textbf{0.106} & 0.072 & 0.158 & \textbf{0.068} & 0.129 \\
ETH-UCY~\cite{Pellegrini2010ECCV,Leal2014CVPR} & 0.346 & 0.464 & 0.318 & 0.426 & 0.444  & 0.670 &  \textbf{0.188}  &  \textbf{0.250}   \\
   
inD~\cite{Bock2020IV}   & 0.079 & 0.099 & 0.068 &  \textbf{0.080} &   0.068 &  \textbf{0.080}   & \textbf{0.064}  & 0.116 \\
JRDB~\cite{Mao2023CVPR} &  0.118 & 0.164 & 0.090 & 0.113 & 0.080 & 0.121 &  \textbf{0.075}  &  \textbf{0.104}  \\
VSCrowd~\cite{Li2022TIP} & 0.428& 0.494 & 0.340 & 0.382 & 0.100 & 0.118
 & \textbf{0.094} & \textbf{0.110}   \\
 FDST~\cite{Fang2019ICME}  & -&- &- &- & 0.050 & 0.094 &\textbf{0.039} & \textbf{0.071}
 \\
croHD~\cite{Sundararaman2021CVPR} &- & - &- &- & 0.039 & 0.049 & \textbf{0.038}  & \textbf{0.046} \\
\bottomrule
    \end{tabular}}
\end{center}
 \label{tab:comparison-using-ground-truth-maps}
\end{table}

\noindent \textbf{Ground Truth Input.} To evaluate the performance of crowd density forecasting models independent of the accuracy of the upstream crowd density estimation module, we conduct experiments using ground truth as input. We train the model with ground truth maps and evaluate the accuracy with them. Since the ground truth 2D positions of the pedestrians are annotated in the above trajectory prediction datasets, we generate ground truth crowd density maps from the pedestrians' positions. We apply a 2D Gaussian filter (with a standard deviation of $3$ px) following the previous approach~\cite{Minoura2021RAL}.

\noindent \textbf{Upstream Perception Modules Input.} To validate the robustness of our method against errors (\eg miss-detection of the pedestrians) in upstream perception modules, we also evaluate the forecasting accuracy in a realistic setting that uses the estimated results from the upstream perception module as inputs in the evaluation phase.  We train the model on ground truth inputs and evaluate the accuracy of the crowd forecasting and trajectory prediction approaches using upstream perception module results. We use the detection and crowd density estimation for crowd density forecasting approaches and the detection\&tracking for trajectory prediction approaches as upstream modules. Following PDFN-ST~\cite{Minoura2021RAL}, we employ UCY in this protocol for upstream detection results input setting. As the \textit{students001} and \textit{uni examples} videos in the UNIV scene are not publicly available, we do not use them in this setting. For the upstream crowd density estimation input setting, we choose the dataset with annotations of pedestrian heads. We use the Faster R-CNN model~\cite{Ren2015NIPS} with FPN~\cite{Lin2017CVPR} as the detector, along with Deep-SORT~\cite{Wojke2017ICIP} as the tracker, pretrained on the MOT17~\cite{Anton2016MOT16} dataset. For the detector, we set the confidence threshold to $0.2$. We utilize the implementation of MMTracking~\cite{mmtrack2020}. We use the STEERER~\cite{Han2023ICCV}, pretrained with the JHU-CROWD++~\cite{Sindagi2022TPAMI}, as a crowd density estimator. 

\subsection{Forecasting Accuracy Comparison} 
\noindent \textbf{Ground Truth Input.} \label{comparision-using-ground-truth-as-inputs}
We compare our model with crowd density forecasting and trajectory prediction models using the ground truth input evaluation protocol. \cref{tab:comparison-using-ground-truth-maps} summarizes the performance of the methods on trajectory prediction and crowd analysis datasets. As shown in \cref{tab:comparison-using-ground-truth-maps}, our approach outperforms previous state-of-the-art method PDFN-ST on all trajectory prediction datasets: SDD, ETH-UCY, inD, and JRDB datasets. For instance, on SDD, we reduce $AD_{JS}$ by $5.6\%$ (from $0.072$ to $0.068$), and reduce $FD_{JS}$ by $18\%$ (from $0.158$ to $0.129$) compared to PDFN-ST. Additionally, our CrowdMAC significantly outperforms the trajectory comparison models. CrowdMAC achieves a substantial gain on SDD, ETH-UCY, inD, and JRDB on $AD_{JS}$ metric compared to the Social-Transmotion. Trajectory prediction methods often assume fully observed agent trajectories, with models trained and evaluated on complete data. However, in the crowd density forecasting evaluation, not all pedestrians are consistently visible due to camera limitations or occlusions, which can reduce trajectory prediction accuracy. As a result, the accuracy gap between trajectory prediction and crowd density forecasting is more pronounced in datasets from surveillance cameras with narrow fields of view (\eg, ETH-UCY and VSCrowd) compared to those from drones with wider views (\eg, SDD and inD).

As shown in \cref{tab:comparison-using-ground-truth-maps}, CrowdMAC also outperforms existing both crowd density forecasting method and trajectory prediction methods on all crowd analysis datasets: FDST, croHD, and VSCrowd datasets on $AD_{JS}$  and  $FD_{JS}$ metric. The performance of the trajectory prediction model on crowd analysis datasets is significantly worse than on trajectory prediction datasets. This discrepancy arises because, unlike trajectory prediction datasets captured from a high and distant bird's-eye view, crowd analysis datasets often include scenes recorded from surveillance cameras that are relatively close to people. Consequently, many individuals appear partway through the observation frames. The trajectory prediction model, which assumes that all pedestrians are present in all observation frames, fails to handle such cases effectively.

\begin{table}[tb]
\caption{Comparison of crowd density forecasting and trajectory prediction approaches using inputs from the upstream detector~\cite{Ren2015NIPS} and tracker~\cite{Wojke2017ICIP} perception modules (see \cref{sec:eval_protoc}).}

\begin{center}
\resizebox{\columnwidth}{!}{
\begin{tabular}{cccccccgg}
\toprule
 \multirow{4}{*}{Dataset} &  \multicolumn{4}{c}{Trajectory Prediction} & \multicolumn{4}{c}{Crowd Density Forecasting}  \\ 
 \cmidrule(l{2pt}r{3pt}){2-5}  \cmidrule(l{2pt}r{3pt}){6-9} 

  &\multicolumn{2}{c}{Y-Net~\cite{Mangalam2021ICCV}}   &\multicolumn{2}{c}{Social-Trans.~\cite{Saadatnejad2024ICLR}} &\multicolumn{2}{c}{PDFN-ST~\cite{Minoura2021RAL}} & \multicolumn{2}{g}{Ours}  \\
 \cmidrule(l{2pt}r{3pt}){2-3}  \cmidrule(l{2pt}r{3pt}){4-5} 
 \cmidrule(l{2pt}r{3pt}){6-7} 
 \cmidrule(l{2pt}r{3pt}){8-9} 
&   $AD_{JS}$ & $FD_{JS}$ & $AD_{JS}$ & $FD_{JS}$ & $AD_{JS}$ & $FD_{JS}$ & $AD_{JS}$ & $FD_{JS}$  \\
\toprule
UNIV ~\cite{Leal2014CVPR} & 0.216 & 0.300 & 0.385 & 0.431 & 0.473 & 0.533  & \textbf{0.281}  & \textbf{0.291}\\
ZARA1~\cite{Leal2014CVPR} & 0.778& 0.809&  0.475 &  0.568 & 0.737 & 1.047 & \textbf{0.385} & \textbf{0.389}  \\ 
ZARA2~\cite{Leal2014CVPR} &  0.882 & 0.889 & 0.463 & 0.645 & 0.554 & 0.621 &  \textbf{0.358} & \textbf{0.417}\\
VSCrowd~\cite{Li2022TIP}  &1.378& 1.974 & 0.726 & 0.807 & 0.250 & 0.389 &  \textbf{0.115} &  \textbf{0.125}   \\

 \bottomrule
\end{tabular}}
\end{center}
 \label{tab:comparison-using-detection-input}
\end{table}

\begin{table}[tb]
\caption{Comparison of crowd density forecasting approaches using inputs from the upstream crowd density estimation~\cite{Han2023ICCV} perception module (see \cref{sec:eval_protoc}). }

\begin{center}
\resizebox{0.8\columnwidth}{!}{
\begin{tabular}{cccgg}
\toprule
 \multirow{2.5}{*}{Dataset} &\multicolumn{2}{c}{PDFN-ST~\cite{Minoura2021RAL}} & \multicolumn{2}{g}{Ours}  \\
 \cmidrule(l{2pt}r{3pt}){2-3} \cmidrule(l{2pt}r{3pt}){4-5}
  & $AD_{JS}$ & $FD_{JS}$ & $AD_{JS}$ & $FD_{JS}$  \\
\toprule
ZARA2~\cite{Leal2014CVPR}
& 0.517 & 0.576 &  \textbf{0.478} &  \textbf{0.479} \\

VSCrowd~\cite{Li2022TIP} & 0.240 & 0.259 & \textbf{0.233} &   \textbf{0.258}  \\
FDST~\cite{Fang2019ICME} & 0.272 & 0.333  & \textbf{0.198} & \textbf{0.200}   \\
 \bottomrule
\end{tabular}}
\end{center}
 \label{tab:comparison-using-crowd-density-estimation-input}
\end{table}

\noindent \textbf{Upstream Perception Modules Input.} \label{comparision-using-upstream-perception-modules-results-as-inputs} We validate the robustness of our method against errors in upstream perception modules using detection results as input, as shown in \cref{tab:comparison-using-detection-input}, and crowd density estimation results as input, as shown in \cref{tab:comparison-using-crowd-density-estimation-input}. Our CrowdMAC consistently outperforms both the crowd density forecasting models and the trajectory prediction models in both settings. For instance, CrowdMAC reduces $AD_{JS}$ by $54\%$ on VSCrowd compared to PDFN-ST with the detection input setting. CrowdMAC reduces $AD_{JS}$ by $7.5\%$ on ZARA2, by $2.9\%$ on VSCrowd, and by $17\%$ on FDST compared to PDFN-ST with the detection input setting. Our CrowdMAC also significantly outperforms trajectory prediction models. All these results confirm that the proposed model significantly improves robustness against real data, which is reasonable as the standard prediction system does not explicitly account for upstream perception errors.

\subsection{Robustness against Miss-detection} 
We explore the robustness of the crowd density forecasting models against the miss-detection of pedestrians with the synthetically-generated and realistic miss-detections.

\begin{figure}[tb]
\centering
\begin{tabular}{cc}
\begin{minipage}{0.45\linewidth}
    \begin{center}
        \includegraphics[clip, width=\hsize]{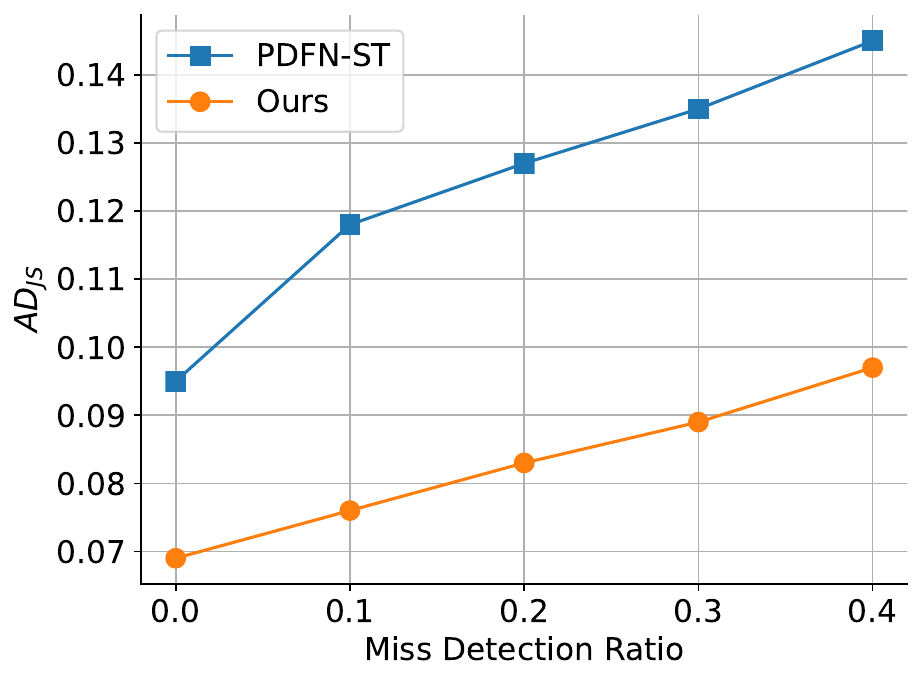}  
        {(a) SDD with synthetic miss-detections.}
    \end{center}
\end{minipage}
&
\begin{minipage}{0.45\linewidth}
    \begin{center}
        \includegraphics[clip, width=\hsize]{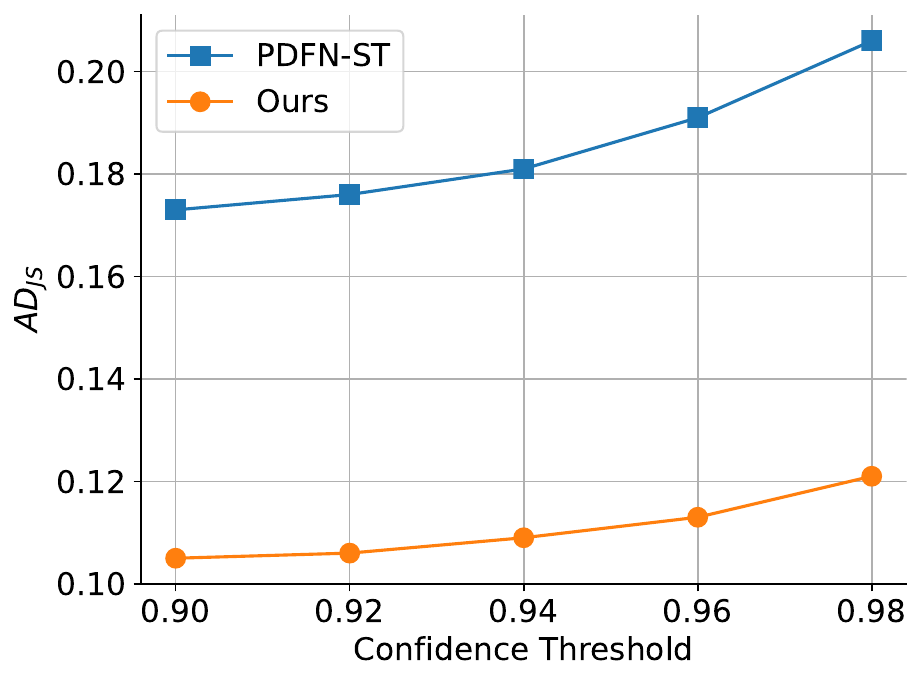}  
        {(b) FDST with realistic miss-detections.}
    \end{center}
\end{minipage}
\end{tabular}
\caption{We compare the robustness of the models with synthetic miss-detections on the SDD and with realistic miss-detections on the FDST.}
 \label{fig:changing-miss-detection-ratio}
\end{figure}

\noindent \textbf{Synthetic Miss-Detection.} 
We evaluate the robustness against synthetically generated miss-detections. We train the models with ground-truth input, and synthetic miss-detections are generated by randomly sampling pedestrians in the frames that will not be used as input during the evaluation. \cref{fig:changing-miss-detection-ratio} (a) summarizes the $AD_{JS}$ metric of the previous approach (PDFN-ST) and CrowdMAC with various miss-detection ratios. Since the performance drop of our model is lower than PDFN-ST among multiple miss-detection ratios, our model is robust against miss-detections compared to the PDFN-ST.

\noindent \textbf{Realistic Miss-Detection.} 
We further investigate the robustness against miss-detection using data preprocessed by the pedestrian detection module~\cite{Ren2015NIPS} on the FDST. We varied the confidence threshold value of the object detector from $0.2$ (as described in \cref{sec:eval_protoc}) to higher values ($>0.9$) to generate input data with miss-detections. Increasing the detection confidence thresholds resulted in more miss-detections of pedestrians (\ie, false negatives in detection). In~\cref{fig:changing-miss-detection-ratio} (b), the $AD_{JS}$ metric of each method (PDFN-ST and Ours) is presented for various detection confidences. Similar to \cref{fig:changing-miss-detection-ratio} (a), our proposed method exhibits a lower performance drop compared to the PDFN-ST, highlighting its robustness against realistic miss-detections. The result of VSCrowd is provided in the supplementary.

\begin{table}[tb]
 \caption{Ablation study of temporal-density-aware masking (TDM) on the SDD and FDST. The TDM consists of a temporal-aware masking (TM) ratio and density-aware masking (DM). }
\begin{center}
\resizebox{0.8\columnwidth}{!}{
\begin{tabular}{cccccc}
\toprule
 \multirow{2.5}{*}{TM Ratio} & \multirow{2.5}{*}{DM}  & \multicolumn{2}{c}{SDD} & \multicolumn{2}{c}{FDST}  \\
\cmidrule(l{2pt}r{3pt}){3-4} \cmidrule(l{2pt}r{3pt}){5-6}  
& & $AD_{JS}$ & $FD_{JS}$ & $AD_{JS}$ & $FD_{JS}$  \\
 \midrule
- & - & 0.072  &  0.132  &  0.046 & 0.082\\
- & \checkmark & 0.070 & 0.130 & 0.043 & 0.075 \\
\checkmark & - &  0.071  & 0.130 & 0.045 & 0.079 \\ 
\rowcolor{Gray}
\checkmark & \checkmark & \textbf{0.068} & \textbf{0.129} &  \textbf{0.039} & \textbf{0.071} \\
\bottomrule
\end{tabular}}
\end{center}
 \label{tab:ablation-tdom}
\end{table}

\subsection{Ablation Studies} 
We perform two ablation studies on the proposed TDM module to assess its effectiveness. In this section, we use the ground truth input evaluation protocol as mentioned in \cref{comparision-using-ground-truth-as-inputs}.

\noindent \textbf{Effect of the Temporal-Density-aware Masking.}
\cref{tab:ablation-tdom} shows the effect of each component (TM, DM), in TDM. It can be seen that both of the components improved the density forecasting performance.

\noindent \textbf{Effect of multi-task masking.}
\cref{tab:ablation-multi-task_masking} shows the effect of mul-task masking. The model trained with a single future prediction task performs considerably worse compared to the model trained with the multi-task approach. In particular, the history prediction task significantly contributes to the improved performance.

\noindent \textbf{Choice of Temporal-aware Masking (TM)
Ratio Function.} 
We explore the design of the TM ratio function in \cref{eq:ratio}. We compare the results with the approaches that employ different masking ratio functions in~\cref{tab:ablation-temporal-aware-masking-function}. The first row shows the results of the approach that uses the constant masking ratio during the training. From the second row to the fifth row show the approaches that employ various masking ratio functions instead of the function in \cref{eq:ratio}, such as concave (square root), linear, convex (square, cubic), respectively. We observe that our temporal-aware masking ratio function outperforms the approach with the fixed masking ratio as well as the approaches that employ the other mask ratio functions. 

\noindent \textbf{Ablation Study of $\tau$ and $\lambda_{max}$}. We conducted ablations with different values of $\tau$ in density-aware observation masking and $\lambda_{max}$ in temporal-aware observation masking, as shown in~\cref{tab:ablation-lambda-tau}. We chose $500$ for $\tau$ and $9$ for $\lambda_{max}$ as the default settings.

\begin{figure}[tb]
\begin{center}
\resizebox{\columnwidth}{!}{
\begin{minipage}{\hsize}
  \begin{center}
    \includegraphics[clip, width=\hsize]{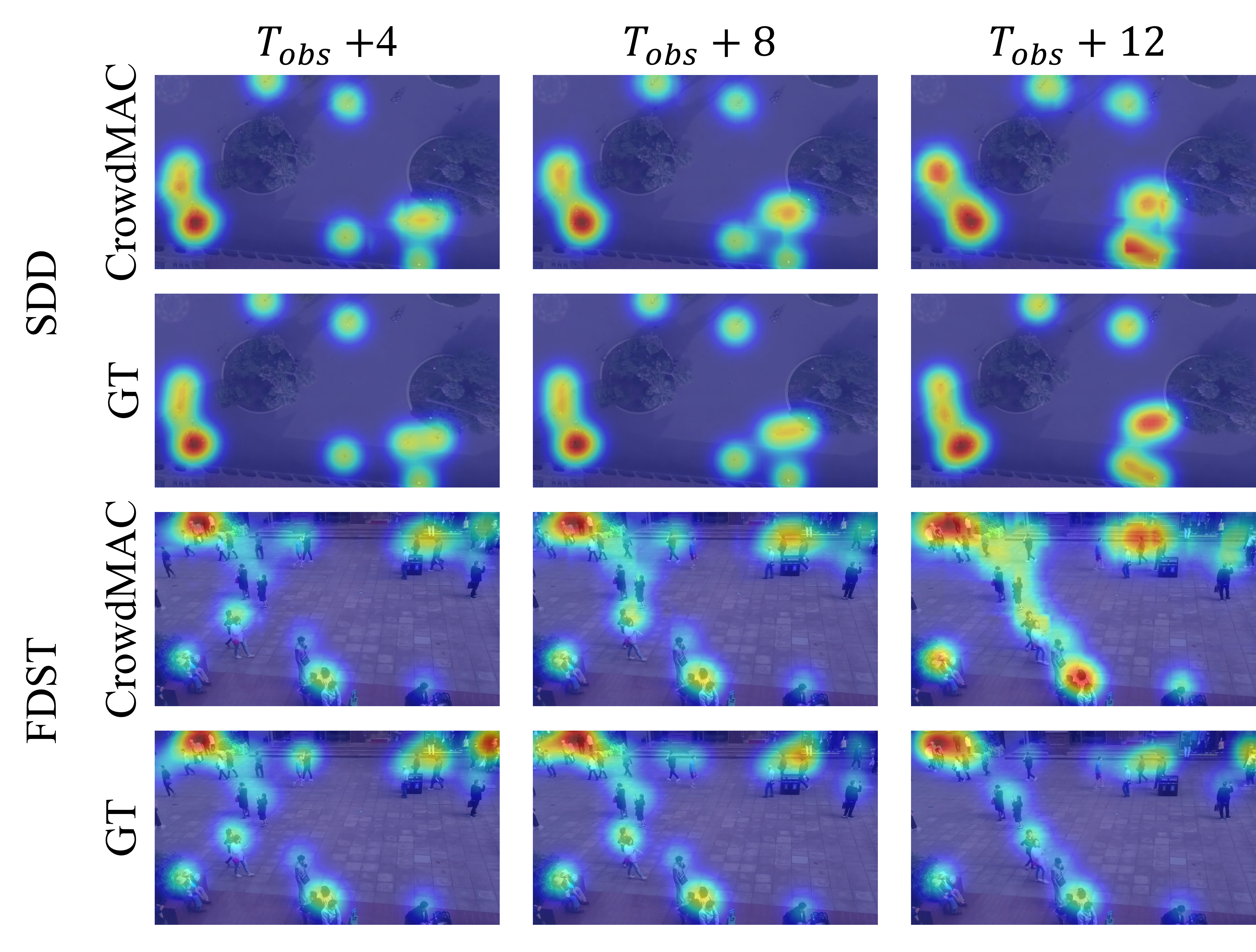} 
  \end{center}
\end{minipage}}
\caption{Qualitative results on the SDD and FDST datasets. The density maps are visualized as a heatmap and overlayed onto the RGB image for visualization purposes.}
\label{fig:qualitative-sdd}
\end{center}
\end{figure}

\begin{table}[tb]
 \caption{Ablation study of multi-task masking on the SDD and FDST.}
\begin{center}
\resizebox{\columnwidth}{!}{
\begin{tabular}{cccccccc}
\toprule
 \multirow{2.5}{*}{Future Pred.} & \multirow{2.5}{*}{Interpolation}  & \multirow{2.5}{*}{Past Pred.}  & \multicolumn{2}{c}{SDD} & \multicolumn{2}{c}{FDST}  \\
 \cmidrule(l{2pt}r{3pt}){4-5} \cmidrule(l{2pt}r{3pt}){6-7}  
 &  &   & $AD_{JS}$ & $FD_{JS}$  & $AD_{JS}$ & $FD_{JS}$ \\
 \midrule
\checkmark  &- & - & 0.080  &  0.146 &  0.054 & 0.089 \\
\checkmark  & \checkmark & - & 0.070 & 0.129 &    0.041 & 0.074 \\
\checkmark  & -  & \checkmark &  0.075  &  0.143 & 0.049 & 0.083\\ 
\rowcolor{Gray}
\checkmark &\checkmark & \checkmark & \textbf{0.068} & \textbf{0.129}  & \textbf{0.039} & \textbf{0.071} \\
\bottomrule
\end{tabular}}
\end{center}
 \label{tab:ablation-multi-task_masking}
\end{table}

\begin{table}[tb]
 \caption{Ablation study of temporal-aware masking (TM) ratio function on the SDD and FDST.}  
\begin{center}
\resizebox{0.8\columnwidth}{!}{
\begin{tabular}{ccccc}
\toprule
\multirow{2.5}{*}{TM  Ratio Function}& \multicolumn{2}{c}{SDD} & \multicolumn{2}{c}{FDST} \\  \cmidrule(l{2pt}r{3pt}){2-3} \cmidrule(l{2pt}r{3pt}){4-5}  
& $AD_{JS}$ & $FD_{JS}$ & $AD_{JS}$ & $FD_{JS}$ \\

\midrule
Constant & 0.077 & 0.136 & 0.041 & 0.074 \\ \hline
Square Root & 0.076 & 0.138 & 0.041 & 0.073  \\
Linear &  0.074& 0.137 &  0.042 & 0.074\\
Square & 0.071 & \textbf{0.129} & \textbf{0.039} & 0.072   \\
Cubic & 0.075 & 0.134 & 0.041 & 0.073\\
\rowcolor{Gray}
Ours & \textbf{0.068} & \textbf{0.129}  &  \textbf{0.039} & \textbf{0.071} \\
\bottomrule
\end{tabular}}
\end{center}
 \label{tab:ablation-temporal-aware-masking-function}
\end{table}

\begin{table}[tb]
 \caption{Ablation Study of $\tau$ and $\lambda_{max}$: We summarize the $AD_{JS}$ metric with different values of $\tau$ and $\lambda_{max}$ on SDD and FDST. We choose $500$ for $\tau$ and $9$ for $\lambda_{max}$ as the default setting.} 
\begin{center}
\resizebox{\columnwidth}{!}{
\begin{tabular}{cc}
{\footnotesize (a) SDD}&{\footnotesize (b) FDST}
\\
\begin{tabular}{c|cccc}
\toprule
 \multirow{2}{*}{$\tau$}  &  \multicolumn{4}{c}{$\lambda_{max}$}  \\
\cline{2-5}
  & 5 & 7 & 9 & 11\\
\cline{1-5}
 300 & 0.075 & 0.071 & 0.069 & 0.069  \\ 
 500 & 0.069 & 0.070 & \textbf{0.068} & 0.069  \\ 
 700 & 0.070 & 0.076 & 0.073 & 0.072 \\ 
\bottomrule
\end{tabular}
&
\begin{tabular}{c|cccc}
\toprule
 \multirow{2}{*}{$\tau$}  &  \multicolumn{4}{c}{$\lambda_{max}$}  \\
\cline{2-5}
  & 5 & 7 & 9 & 11\\
\cline{1-5}
 300 & 0.041  & 0.043  & 0.042 & 0.053   \\ 
 500 & 0.041 & 0.041 & \textbf{0.039}   & 0.053 \\ 
 700 & 0.040  & 0.042 & 0.041 &  0.054\\ 
\bottomrule
\end{tabular}
\end{tabular}}
\end{center}
 \label{tab:ablation-lambda-tau}
\end{table}

\subsection{Qualitative Evaluation}
\cref{fig:qualitative-sdd} visualizes the qualitative results of the proposed method as well as the ground-truth density maps on the SDD and FDST datasets. The proposed method accurately forecasts the crowd density maps. In the supplementary material, we also provide video visualization and the visualized comparison with the PDFN-ST.

\section{Conclusion}
In summary, we propose CrowdMAC, a new learning framework for crowd density forecasting, which operates in a masked completion fashion. CrowdMAC is simultaneously trained to forecast future crowd density maps from masked past observation maps and reconstruct the masked observation maps to improve the robustness of the crowd density forecasting task against the miss-detection of pedestrians. We further propose Temporal-Density-aware Masking (TDM) that non-uniformly masks the patches in the observed crowd density maps, facilitating training and obtaining better forecasting performance. Moreover, we introduce multi-task masking to enhance the efficiency in training. Empirical results show that our method outperforms the state-of-the-art crowd density forecasting methods, as well as trajectory prediction approaches, and achieves robustness against synthetically generated or realistically occurring miss-detections.

{\small
\bibliographystyle{ieee_fullname}
\bibliography{refs}
}

\end{document}